\documentclass[10pt,letterpaper,twocolumn]{article}
\usepackage[latin9]{inputenc}
\usepackage{array}
\usepackage{booktabs}
\usepackage{bbding}
\usepackage{url}
\usepackage{multirow}
\usepackage{amsmath}
\usepackage{amssymb}
\usepackage{graphicx}

\makeatletter

\pdfpageheight\paperheight
\pdfpagewidth\paperwidth

\providecommand{\tabularnewline}{\\}

\usepackage{iccv}
\usepackage{times}
\usepackage{epsfig}
\usepackage{graphicx}
\usepackage{bbm}

 \iccvfinalcopy 

\def\iccvPaperID{3598} 
\def\httilde{\mbox{\tt\raisebox{-.5ex}{\symbol{126}}}}

\ificcvfinal\fi

\@ifundefined{showcaptionsetup}{}{%
 \PassOptionsToPackage{caption=false}{subfig}}
\usepackage{subfig}
\makeatother
\usepackage{authblk}
\begin{document}

\title{}

\title{High-Level Perceptual Similarity is Enabled by Learning Diverse Tasks}

\author[1,2]{Amir Rosenfeld}
\author[2]{Richard Zemel}
\author[1]{John K. Tsotsos}

\affil[1]{York University}
\affil[2]{University of Toronto}
\affil[1,2]{Toronto, Canada}

\maketitle

\begin{abstract}
Predicting human perceptual similarity is a challenging subject of
ongoing research. The visual process underlying this aspect of human
vision is thought to employ multiple different levels of visual analysis
(shapes, objects, texture, layout, color, etc). In this paper, we
postulate that the perception of image similarity is not an explicitly
learned capability, but rather one that is a byproduct of learning
others. This claim is supported by leveraging representations learned
from a diverse set of visual tasks and using them jointly to predict
perceptual similarity. This is done via simple feature concatenation,
without any further learning. Nevertheless, experiments performed
on the challenging Totally-Looks-Like (TLL) benchmark significantly
surpass recent baselines, closing much of the reported gap towards
prediction of human perceptual similarity. We provide an analysis
of these results and discuss them in a broader context of emergent
visual capabilities and their implications on the course of machine-vision
research.
\end{abstract}

\section{Introduction}

The many capabilities exhibited by human visual perception entail
multiple aspects of image analysis, not all of which are fully understood.
Of these capabilities, some are not explicitly supervised -- though
the current trend in the vision community is to learn them in a data-driven
manner. Examples include saliency \cite{koch1985shifts} and object
tracking \cite{moore1978visual}. Others such as object naming are
somewhat supervised (though the extent of supervision and its quality
are varied and debatable). In contrast, successful methods in machine
learning seem to be mostly supervised with the an explicit goal in
mind: to do well on a certain (increasingly large) benchmark, on which
they are trained (\cite{lin2014microsoft,russakovsky2015imagenet,alp2018densepose,krishna2017visual,OpenImages}).
Unfortunately, training data is not easy to produce when the expected
output goes beyond image-level labels, such as dense keypoint annotations
\cite{alp2018densepose} or precise segmentation masks \cite{lin2014microsoft},
often requiring laborious human annotations and careful instructions
and definitions. We claim that in some cases, if a task can be already
achieved as a by-product of succeeding in a variety of others, attempting
to learn it explicitly may not be the most effective solution.

This paper exemplifies our claim through the task of human perceptual
similarity (HPS): the human judgment of whether two images are similar
or not. While recent works have empirically shown that deep-learned
representations perform this task with high accuracy \cite{battleday2017modeling,10.3389/fpsyg.2017.01726,peterson2016adapting,zhang2018unreasonable},
others have shown the contrary. Specifically, \cite{rosenfeld2018totally}
experiments on the Totally-Looks-Like (TLL), where deep-learned representations
perform quite poorly indeed.

\begin{figure}
\includegraphics[width=1\columnwidth]{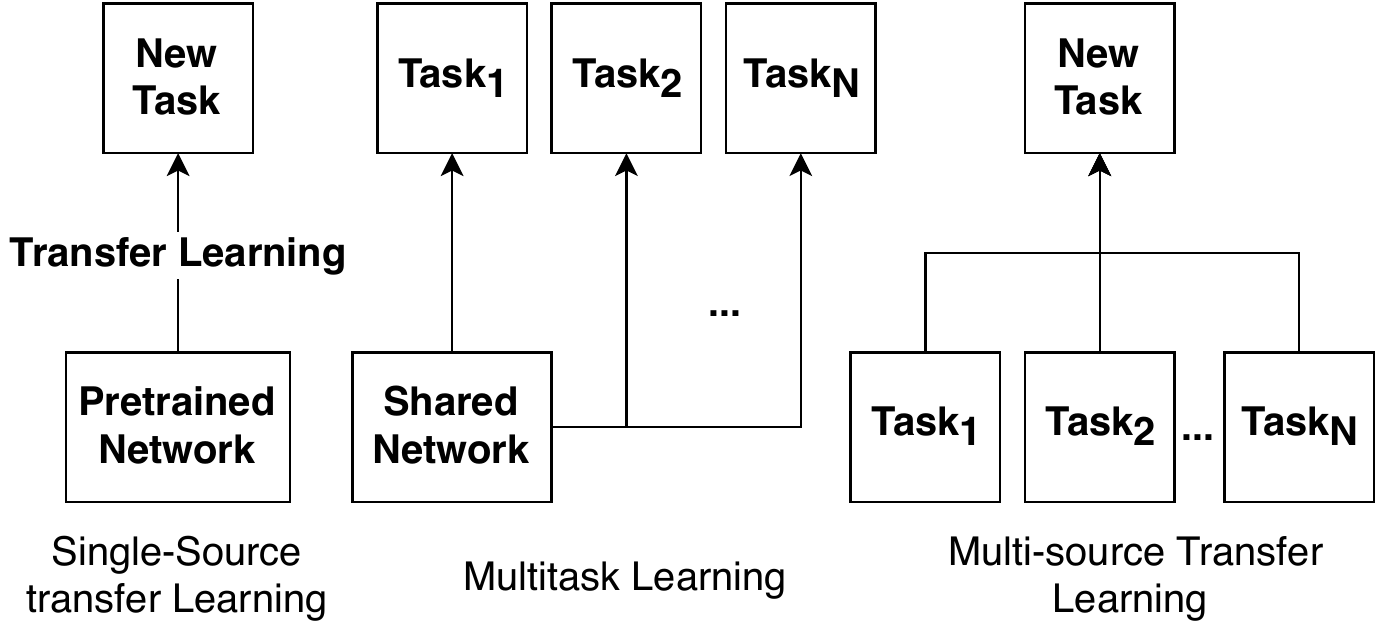}

\caption{\label{fig:multisource}Types of multi-task learning: transfer learning
utilizes a single source to adapt to a new task. Multitask learning
learns all tasks simultaneously. We suggest leveraging\emph{ multiple
}sources for a single task. In this work, we show the merits of this
method though the source representations are only used without any
fine-tuning.}
\end{figure}

We propose to approach HPS prediction by leveraging representations
learned from a variety of visual tasks. Intuitively, a diverse set
of tasks will be better at covering the broad set of features thought
to underlie perceptual similarity. We utilize image retrieval \cite{radenovic2018fine},
semantic keypoint matching \cite{rocco2018end}, object categorization
\cite{russakovsky2015imagenet}, and others. These are tasks for which
obtaining annotations on sufficient amounts of data is manageable,
either by manual annotation or by devising methods to train in a semi-supervised
regime. In contrast, collecting data for perceptual similarity is
more difficult because it may require a quadratic number of labels,
one for each image pair. Exceptions are cases where image pairs are
generated automatically \cite{zhang2018unreasonable}, which we claim
limits their diversity, or if images can be grouped into equivalence
classes, but then the problem is reduced to classification or learning
an embedding space \cite{schroff2015facenet} to reflect desired inter-
and intra-class distances. Our approach is different from common practices;
in transfer-learning a representation learned from a source task is
tuned to perform well on a target task. Implicitly, this assumes that
the source and target tasks are sufficiently related \cite{zamir2018taskonomy}.
In multi-task learning a single architecture is trained jointly on
multiple tasks with some amount of shared representation \cite{ruder2017overview},
in hope that the resultant net will either be more compact or benefit
positively from joint training. We propose the opposite: a single
task can benefit from using a variety of specialized features learned
on others. This is reminiscent of Multiple-Kernel-Learning \cite{gonen2011multiple}
which was popular before the current deep learning era. These differences
are illustrated in Figure \ref{fig:multisource}.

We make the following contributions:
\begin{enumerate}
\item We propose a simple but effective practice of leveraging multiple
specialized learned representations where it is unlikely to obtain
sufficient supervision.
\item Experimentally, we show this approach yields significantly improved
performance on the TLL dataset.
\item We provide an analysis of the contributions of different representations;
in which ways they reliably predict HPS and in which ways they are
still lacking.
\end{enumerate}

\section{Related Work}

\begin{figure}
\subfloat[]{\includegraphics[width=0.33\columnwidth]{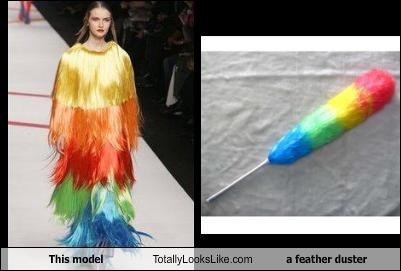}\,\includegraphics[width=0.33\columnwidth]{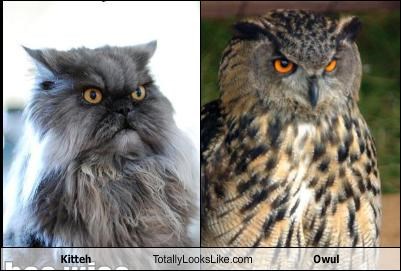}\,\includegraphics[width=0.33\columnwidth]{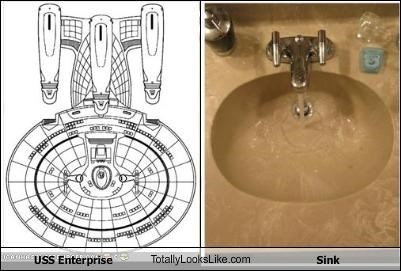}

}
\begin{centering}
\subfloat[]{\begin{centering}
\includegraphics[width=0.2\columnwidth]{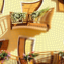}\quad{}\includegraphics[width=0.2\columnwidth]{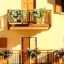}\quad{}\includegraphics[width=0.2\columnwidth]{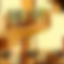}
\par\end{centering}
}
\par\end{centering}
\caption{\label{fig:The-Totally-Looks-Like-dataset.} Top: the Totally-Looks-Like
dataset \cite{rosenfeld2018totally}. The images within each pair
have been judged by some user to be similar. Bottom : images from
\cite{zhang2018unreasonable}, where a user must choose which of the
two distorted versions (left, right) of an image is more similar to
a reference (center). }
\end{figure}

\textbf{Perceptual Similarity}: classical works on perceptual similarity
already recognize it as a multi-faceted \cite{medin1993respects,gati1984weighting,tversky1977features},
knowledge and context dependent \cite{murphy1985role,chi1981categorization}
problem. More recent benchmarks include subjective image quality assessment
with a reference image, which have been serving for evaluating similarity
metrics \cite{sheikh2006statistical,ponomarenko2015image,larson2010most}.
The large-scale BAPPS dataset has been recently introduced by \cite{zhang2018unreasonable},
more geared towards perceptual similarity than quality assessment
per-se. Several lines of work have claimed that human perceptual similarity
judgment is solved to a good extend by CNN-based methods \cite{battleday2017modeling,10.3389/fpsyg.2017.01726,peterson2016adapting}.

\textbf{Emergent properties}: Some visual capabilities such as gaze
direction prediction and hand detection can be explained by using
causal reasoning together with innate capabilities (\eg face detection)
\cite{ullman2012simple}. Object naming does receive some amount of
supervision during childhood and this is indeed shown to assist development
in early stages \cite{booth2002object}. Nevertheless, many other
abilities are either seldom or even never supervised. Examples include
stereopsis, contour integration, perception of motion and other \cite{siu2018development}.
Behavioral patterns linked with vision also emerge; a notable example
is saliency, \ie the prediction of gaze given a visual stimulus \cite{koch1985shifts}.
Though saliency is clearly a measurement of a single behavioral aspect
of the visual system, virtually all recent leading methods of predicting
saliency have been based on purely data-driven methods \cite{bylinskii2015saliency}.
With some rare exceptions \cite{adeli2018learning}, most methods
treat saliency as a goal rather than observing it as a part of a functioning
system. There are are few recent papers that report emergence of useful
visual representations, such as emergence of visual tracking by the
need to color videos in a consistent manner \cite{vondrick2018tracking}
as well as motor \cite{heess2017emergence} or visual \cite{haber2018emergence}
skills.

\textbf{Transfer/Multitask Learning}: transfer learning has already
been established as the tool to enable learning of new tasks by leveraging
already learned ones \cite{yosinski2014transferable}. The transferability
of tasks to related ones has also been explored, \cite{zamir2018taskonomy}.
Recently, the work of \cite{1902.03545} has shown a method of predicting
which feature extractors will perform well on a given task. In multi-task
learning, a single network is adapted to multiple tasks \cite{ruder2017overview}.
The shared representation is more compact than using an exclusive
network for each task independently. We propose not to adapt one net
to multiple representations, but to adapt multiple representations
to a single task. Related approaches exist in NLP where pre-trained
representations via multiple tasks turn out useful for many downstream
ones \cite{devlin2018bert,kiros2015skip}.

\section{Approach}

Our goal is to predict human perceptual similarity. We briefly define
the setting in \cite{rosenfeld2018totally} who approach it as image
retrieval. A set of $N$ images $(L_{i},R_{i}),i\in[1\dots N]$ is
given, where for each $i$ the image (left) $L_{i}$ was deemed by
a human to be similar to the (right) image$R_{i}$.

\paragraph{Ranking by Similarity\label{par:Ranking-by-Similarity}}

A solution is defined by a similarity matrix $\Phi$ over the pairs:
\begin{equation}
\Phi(L_{i},R_{j})\rightarrow\mathcal{R}
\end{equation}
 computed over each pair of images $i,j$. Given an image $L_{i}$,
and $j\neq k$, $\Phi$ induces a ranking $\prec_{\Phi}$ such that
\begin{equation}
\Phi(L_{i},R_{k})>\Phi(L_{i},R_{j})\Leftrightarrow R_{k}\prec_{\Phi}R_{j}
\end{equation}

In other words, a higher similarity means a lower ranking (where rank
0 means maximally similar). Given this ranking, the quality of $\Phi$
is measured by recall $@1$ : the number of times (out of $N$) where
the first ranked right image matched the original left one: $R_{i}\prec_{\Phi}R_{j}$for
each $j\neq i$.

In all cases, to obtain the similarity matrix $\Phi$ we extract from
each image $I$ some intermediate output of a convolutional neural
net. Denote this output by $F(I)$. Then $\Phi_{i,j}$ is simply the
cosine similarity between the two features:
\begin{equation}
\Phi_{i,j}=\frac{F(L_{i})\cdot F(R_{j})}{\left\Vert F(L_{i})\right\Vert \left\Vert F(R_{j})\right\Vert }
\end{equation}

\paragraph{Combining Representations}

If we have $M$ similarity matrices $\Phi^{m}$ ($m\in[1\dots M]$
denoting the index of a specific similarity type) we obtain a joint
similarity matrix $\Phi_{s}$ from a subset $S\subseteq[1\dots M]$
by simply summing them:

\begin{equation}
\Phi_{s}=\sum_{k\in S}\Phi^{k}\label{eq:combination}
\end{equation}

Alternatively, we can sum a normalized version of each

\begin{equation}
\Phi_{S}=\sum_{k\in S}(\Phi^{k}-\mu_{k})/\sigma_{k}\label{eq:combination-with-weights}
\end{equation}

Where $\mu_{k},\sigma_{k}$ are the estimated mean and standard deviation
of the values of $\Phi^{k}$.

\section{Experiments \& Analysis}

We first describe the TLL data. We then move on to list the various
feature representations used, followed by specifics of image pre-processing
and feature extraction. Finally, we report results with added analysis.

\subsection{Data}

The images of the Totally-Looks-Like (TLL) dataset \cite{rosenfeld2018totally}
were obtained from a popular website \footnote{\url{https://memebase.cheezburger.com/totallylookslike}}
where users are free to upload a pair of images that they somehow
deem similar. It contains overall 6,016 image pairs. Note that this
there is no strict definition or rules of image similarity and uploading
images is fully voluntary. This causes an interesting diversity of
image-pairs, as is evident in the top part of Figure \ref{fig:The-Totally-Looks-Like-dataset.}.
To verify that the image pairs are not uploaded arbitrarily, \cite{rosenfeld2018totally}
performed human experiments, which showed that selected human image
pairings (in a 5AFC test) were both highly consistent within users
and with the originally collected data. For comparison, the bottom
of Figure \ref{fig:The-Totally-Looks-Like-dataset.} shows the setting
in the experiments of Zhang \etal \cite{zhang2018unreasonable}.
In their work, a 2AFC test is made such that a user (or computed similarity
metric) must choose which of two distortions of an image patch is
closer to the reference (undistorted) patch.

\subsection{Feature Representations\label{subsec:Feature-Representations}}

We used the publicly available methods (modifying the architecture
as necessary) for an overall of 11 features representations: \textbf{conv}
- Densenet121 \cite{huang2016densely}, trained on ImageNet \cite{russakovsky2015imagenet};
\textbf{ret }- features for image retrieval \cite{radenovic2018fine}
and \textbf{ret\_w}, which uses the learned whitening by \cite{radenovic2018fine}.
\textbf{shp }- shape-based features for image retrieval \cite{radenovic2018deep};
\textbf{weaka} - weakly supervised alignment of instances \cite{rocco2018end};
\textbf{pl18}, \textbf{pl50} - scene recognition \cite{zhou2017places};
\textbf{pose} - human pose estimation \footnote{implementation of\url{https://github.com/DavexPro/pytorch-pose-estimation}};\textbf{
texture} - classification features with reduced texture sensitivity
($\ie$ shape biased) \cite{geirhos2018imagenet}; \textbf{texture\_SI}
- same as previous but trained on both the modified ImageNet and original;\textbf{
texture\_SII} -same as previous but fine-tuned on original ImageNet
(see \cite{geirhos2018imagenet} for details). The dimensionality
of each representation is between 512 and 2048.

\textbf{Image Preprocessing: }as each left-right image pair $(L_{i},R_{i})$
in TLL may contain text, we first crop the bottom of the images, consistent
with the practice in \cite{rosenfeld2018totally}. Each image ($\ie$
$L_{i}$ and $R_{i}$ ) is then normalized. As in \cite{rosenfeld2018totally},
the shorter side of the image is first resized to 256 and a centered
crop of 224x224 pixels is extracted. All images are standardized by
subtracting the mean and dividing by the per-channel standard deviation
for imagenet-trained networks, $\mu=$ $(0.485,0.456,0.406),$ $\sigma=(0.229,0.224,0.225).$

\textbf{Feature extraction}: most features vector are extracted from
the penultimate layer of a neural network by mean-average pooling
of a window size to match that layer, \ie producing a 1D vector.
Some networks already produce such a vector, such as that of \cite{radenovic2018fine},
hence there is no need to add an extra pooling stage. The vectors
are l-2 normalized.

\subsection{Testing Representations in Isolation}

As described in Section \ref{par:Ranking-by-Similarity} we test the
retrieval performance of each representation. We begin by testing
each in isolation and show the results in Figure \ref{fig:Performance-of-each}.
The results are reported as the number of recalled images out of $N=6,016$.
The ordering of the few leading methods seem to follow a few trends.
The following are qualitative explanations on an intuitive level.
The exact ranking is probably due to other elements that we do not
account for such as the specifics of the training of each method,
data, etc.

\paragraph{Alignment}

The highest ranked methods are generally geared toward alignment/retrieval
of images (\textbf{ret\_w},\textbf{weaka}) and the worst seem to have
little to do with this task (\textbf{pose}). However other results
do not follow this trend, such as the \textbf{shp} results which are
trained retrieving images via shape (contour) features \cite{radenovic2018deep}.

\paragraph{Level of Abstraction}

Specializing networks to be increasingly invariant towards low-level
features makes them in\emph{ }general\emph{ worse }at this task: the
leading \textbf{ret\_w }from a a state-of-the-art image retrieval
method \cite{radenovic2018fine}. Learned by using structure-from-motion
methods on a large-scale dataset \cite{schonberger2015single} to
guide the selection training data. By construction, this method learned
to match low-level features depicting the same structure under different
imaging (viewpoint, lighting) conditions. Note that \textbf{ret }in
the fourth place is the same as \textbf{ret\_w }except the learned
feature-whitening stage. It is worse for retrieval, as reported by
\cite{radenovic2018fine}, consistent with it being less successful
in the TLL task. The second-best is \textbf{weaka} \cite{rocco2018end}
which is also geared at matching semantic keypoints belonging to different
instances of the same category. In other words, their method has learned
to ignore intra-class appearance variations of semantic parts since
they are nuisance factors. The third method, \textbf{conv}, involves
features from a network trained on classification. Such networks are
meant to be invariant to image features that do not indicate the identity
of a category, hence ignoring view-point and intra-category appearance
changes.

Ranks 5-7 (\textbf{texture\_SI},\textbf{ texture\_SII},\textbf{ texture})
are all derived from the shape-biased method \cite{geirhos2018imagenet}.
These trained to be invariant to texture by applying a style-transfer
method \cite{huang2017arbitrary} to training images, hence depicting
the same concept with even more appearances than the original. Apparently
the increased invariance is detrimental for the retrieval process.

Ranks 8-9 (\textbf{pl50},\textbf{pl18}) stem from place-recognition
networks \cite{zhou2017places}. Again, sensitivity for semantic elements
and increased invariance to low to mid-level features may explain
these. However, it is rather surprising that the gap between this
method and categorization-based methods \cite{huang2016densely} is
so large.

The last two methods are expected to be useful: \textbf{shp }\cite{radenovic2018deep}
is a leading method for sketch-based image retrieval, that first transforms
images into an edge-map \cite{dollar2013structured} so that subsequent
features are shape-based. \textbf{pose }\cite{cao2017realtime} also
plays a role in some of the TLL images.

\subsection{Testing Combined Representations\label{subsec:Testing-Combined-Representations}}

We now show the effect of combining multiple specialized representations.
Recall that we denote by $\Phi_{S}$ the combined similarity matrix
obtained by summing all similarity matrices in the subset $S\subseteq[1\dots M]$
(Eq. \ref{eq:combination}). We systematically try all of the non-empty
subsets $S$. This is not computationally demanding for a dataset
of this size, as we can simply store each similarity matrix $\Phi^{m}\in\mathcal{R}^{N\times N}$.
The resultant $\Phi_{S}$ allows us to calculate the recall measure. 

As we have $M=11$ representation types, we test an overall of $2^{11}-1=2047$
combinations. We first calculate maximally achievable result for a
set of size $k$ for $k\in[1\dots M]$.

\begin{figure}
\begin{centering}
\subfloat[\label{fig:Performance-of-each}]{\includegraphics[width=1\columnwidth]{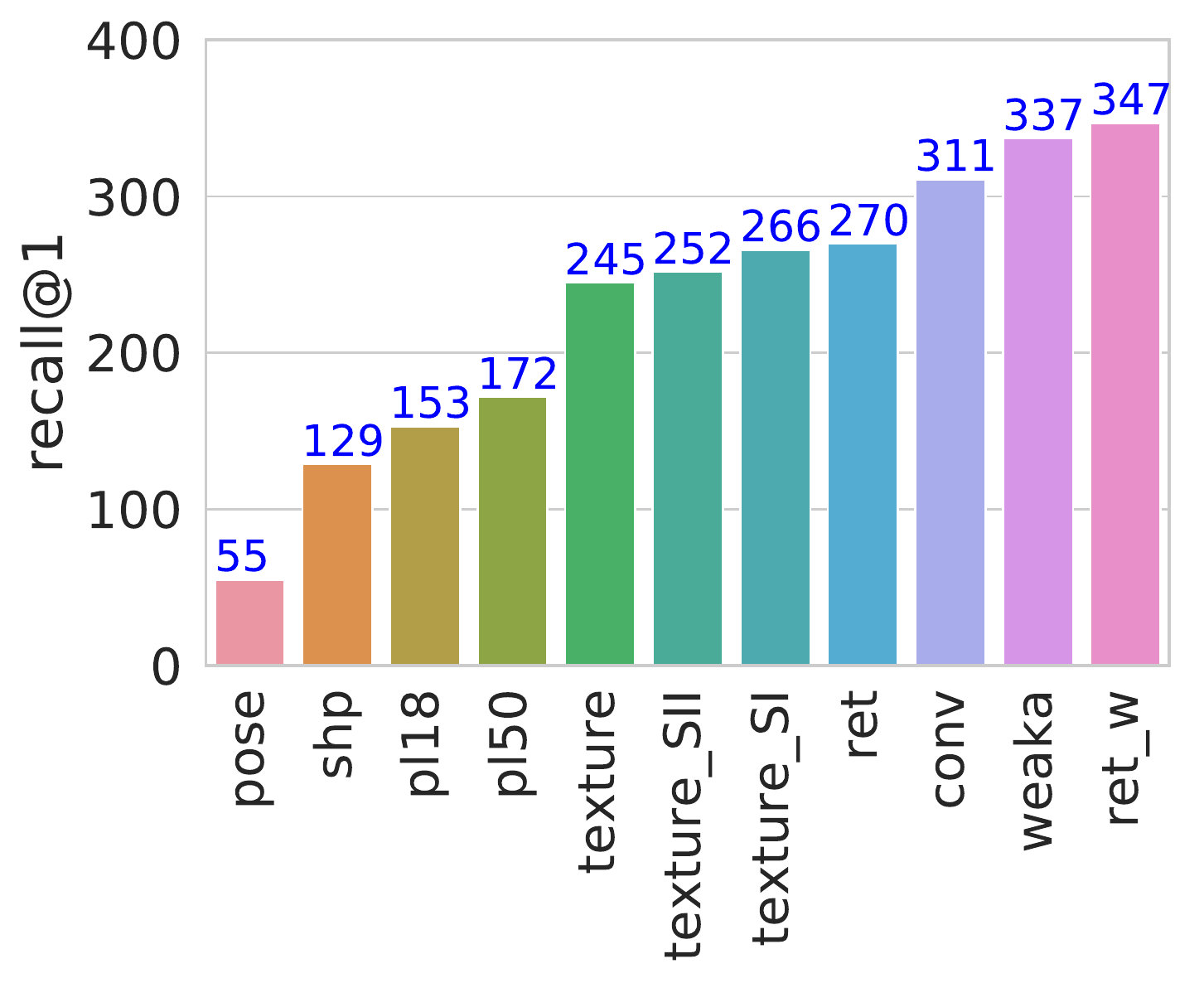}

}
\par\end{centering}
\begin{centering}
\subfloat[\label{fig:Performance-vs.-number}]{\includegraphics[width=1\columnwidth]{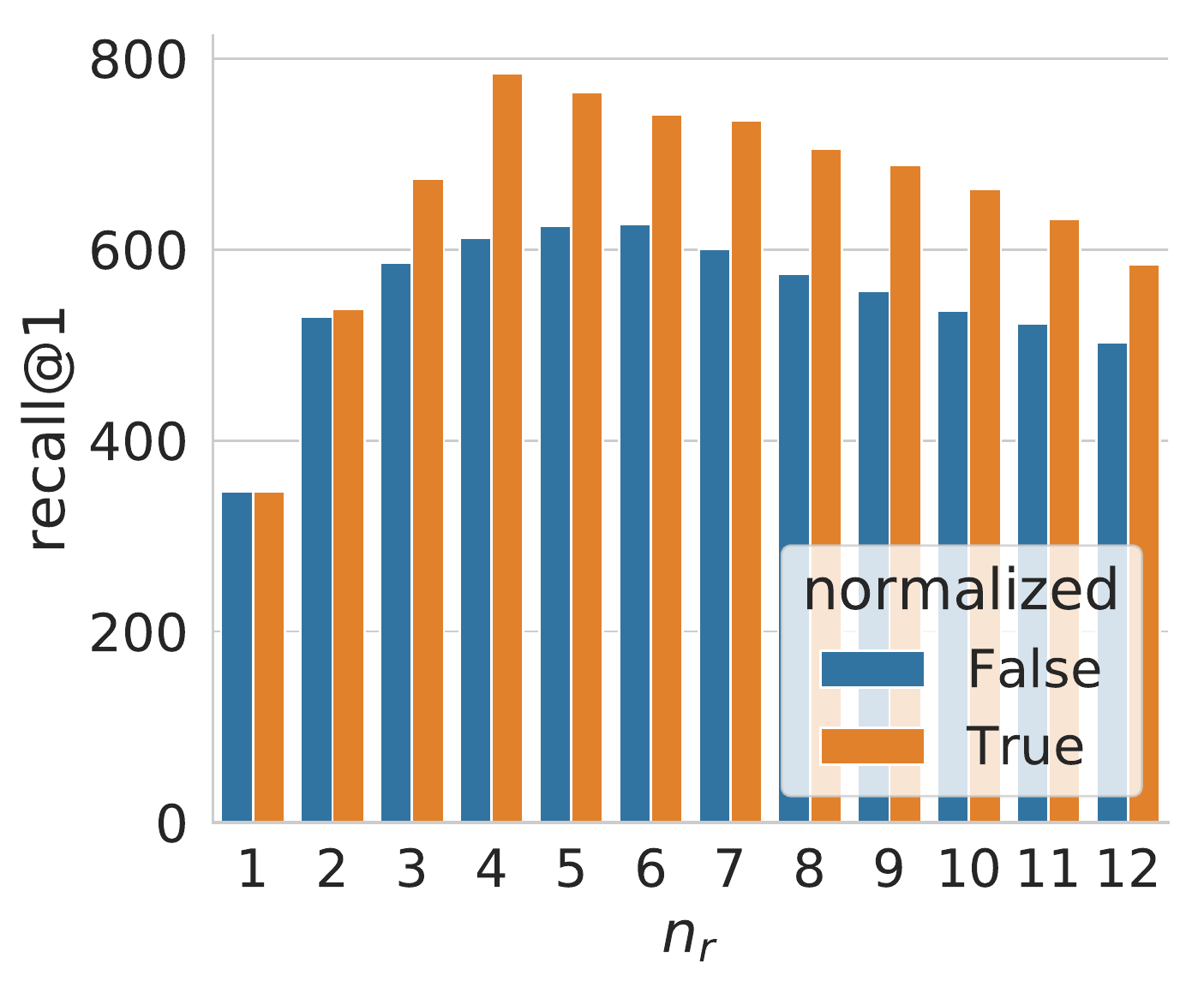}

}
\par\end{centering}
\caption{(a) Performance of each feature in isolation. (b) Performance vs.
number of different representations. Any combination of two representations
easily surpasses the performance of any single one.}
\end{figure}

Figure \ref{fig:Performance-vs.-number} shows the result. We denote
by $n_{r}$ the number of used representations. Evidently, it is always
better to combine more than one representation for this task. But
naively using all of them does not yield the best results; in fact
the unnormalized combination peaks at $n_{r}=6$ and the normalized
combination peaks at $n_{r}=4$. We also see that normalizing the
results to have roughly the same dynamic range (Eq. \ref{eq:combination-with-weights})
yield significantly improved performance. The best recall for an un-normalized
combination is 627, and that of the best-normalized one is 785. The
best result is 2.25x the best single representation (\textbf{ret\_w})
and 2.5x the baseline of \cite{rosenfeld2018totally} (\textbf{conv}).\textbf{ }

In the normalized case it suffices to use combinations of size $n_{r}=4$.
This still leaves us with $\binom{11}{4}=330$ different combinations
to test. We use two methods to assess the utility of each feature
within the combination:

\paragraph{Participation in Leading Combinations}

\begin{figure*}
\subfloat[]{\includegraphics[width=0.31\textwidth]{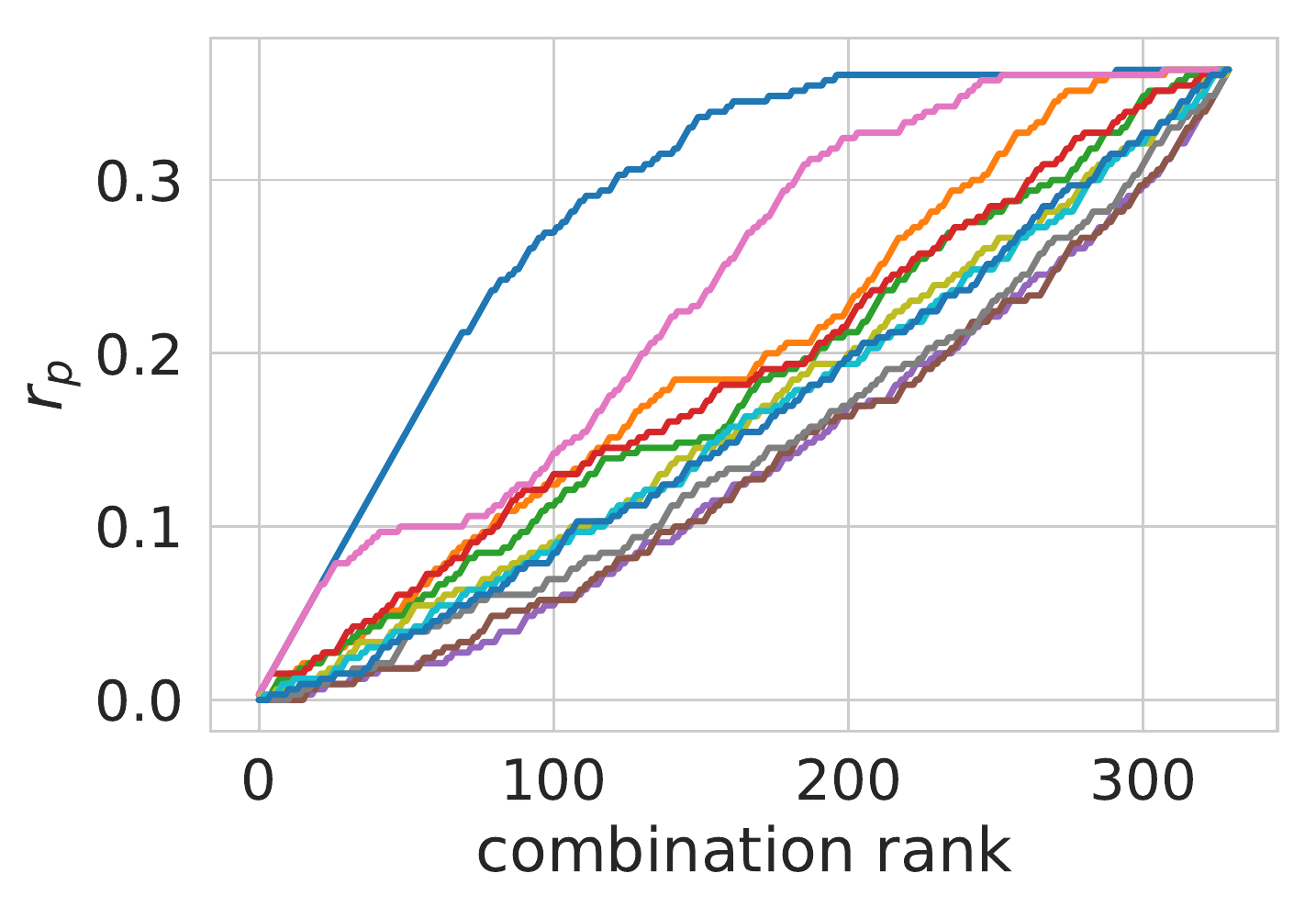}

}\subfloat[]{\includegraphics[width=0.32\textwidth]{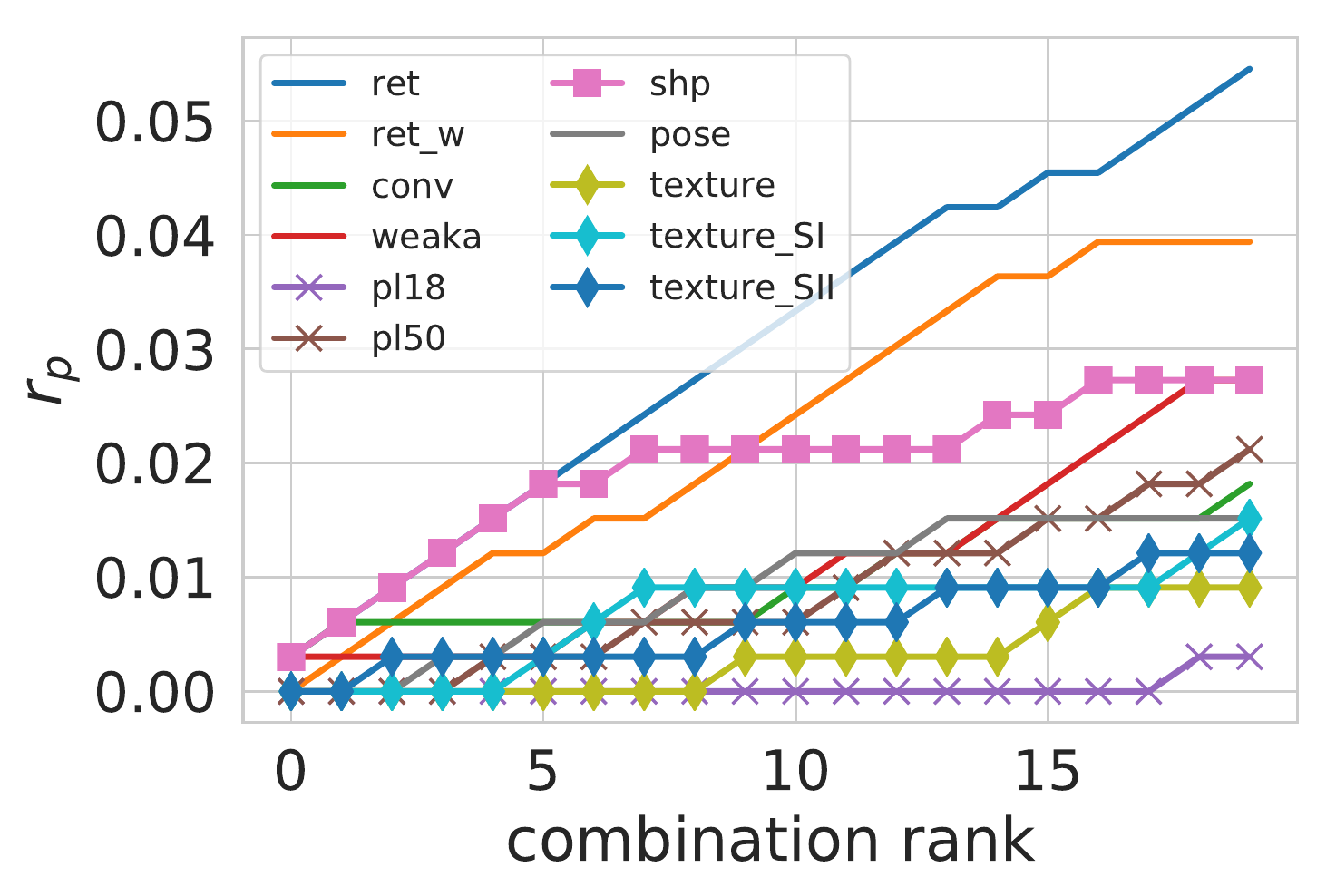}

}\subfloat[]{\includegraphics[width=0.35\textwidth]{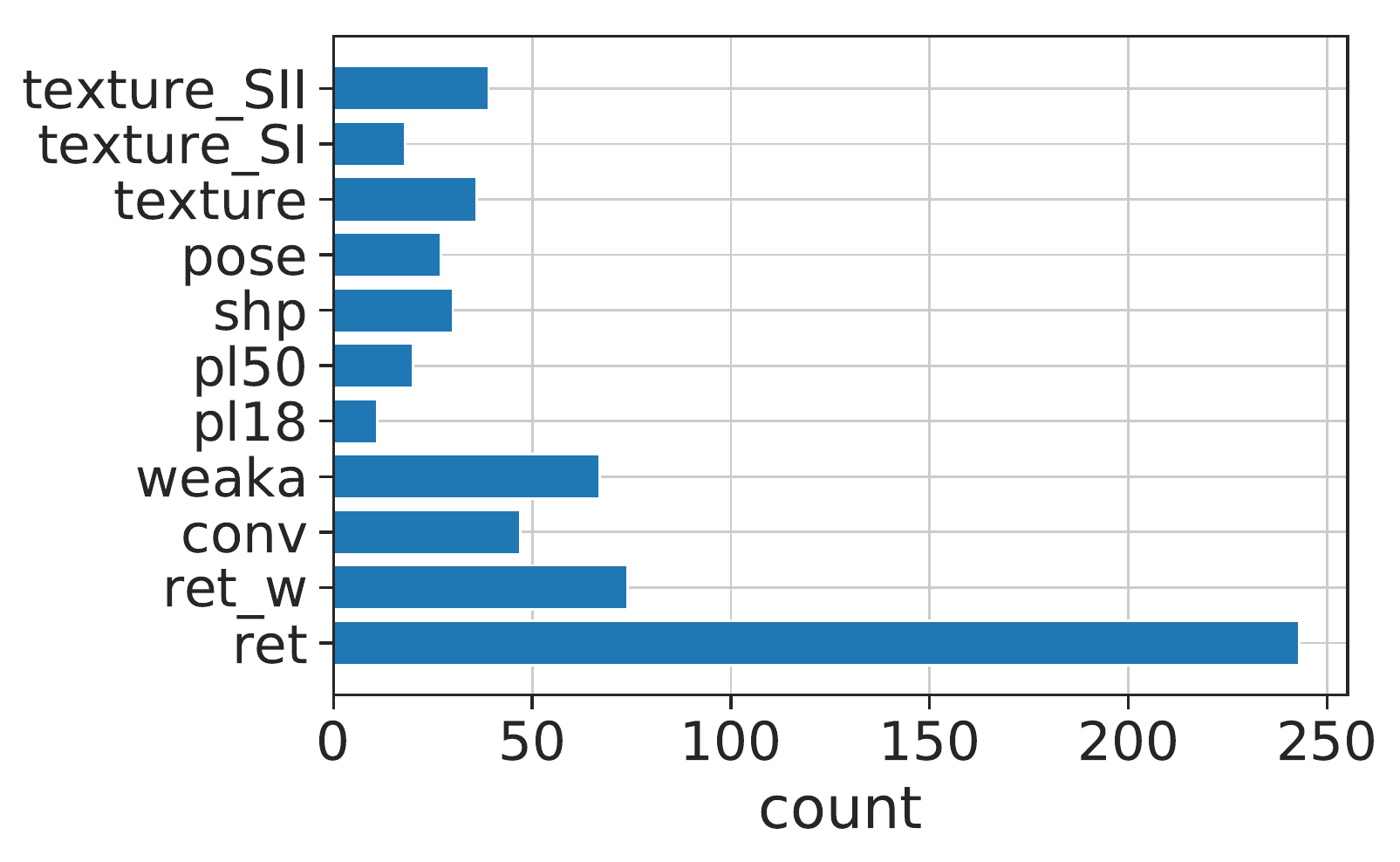}

}

\caption{\label{fig:Feature-importance-vs.}Feature importance (participation
ratio, $r_{p}$) vs. ranking. (a) We plot for each feature the number
of times (y-axis) it appeared in the top-k leading combinations (x-axis).
(b) zoom in on ranks 1-15 of (a). Single feature performance does
not correlate well with its performance in combined with other features
(see text for details). (c) unique feature contributions - number
of images that were correctly retrieved only by a specific feature
but not by others.}

\end{figure*}

First, we look at all 4-feature combinations. For each combination,
we look at the best attainable combination without that feature. For
first $q$ performing combinations (in descending order) we count
the number times a feature $f$ has participated, where $f$ is one
of the feature representations (Section \ref{subsec:Feature-Representations}).
In other words, we calculate for each feature its ratio of participation
$r_{p}$ in leading combinations. This is plotted in Figure \ref{fig:Feature-importance-vs.}(a,b).
We note the weak correspondence between the single-feature performance
(Figure \ref{fig:Performance-of-each}) and this data: the shape (\textbf{shp})
features were almost the worse performers on their own. However, it
seems that except for \textbf{ret}, they are now the leading in terms
of $r_{p}$. This is further accentuated in the next experiment.

\paragraph{Ablating Features}

The leading feature combination involves \textbf{ret},\textbf{ shp},\textbf{
weaka},\textbf{ }and\textbf{ conv}, with a recall of 785. We test
the effect of removing each of this feature type. This is summarized
in Table \ref{tab:Single-representation-contributi}. We find it interesting
that while \textbf{shp }alone achieved a recall of only 129, removing
it from the best combination causes a reduction of 242.

\begin{table}
\noindent\resizebox{\columnwidth}{!}{%
\begin{tabular}{rrrrr>{\raggedleft}p{1.2cm}rr}
\toprule 
ret  & shp  & weaka  & conv  & ret\_w & texture  & recall@1  & $\Delta$\tabularnewline
\midrule 
  & \Checkmark{}  & \Checkmark{}  & \Checkmark{}  & \Checkmark{}  &   & 543  & 242\tabularnewline
\Checkmark{}  &   & \Checkmark{}  & \Checkmark{}  & \Checkmark{}  &   & 638  & 147\tabularnewline
\Checkmark{}  & \Checkmark{}  &   & \Checkmark{}  & \Checkmark{}  &   & 717  & 68\tabularnewline
\Checkmark{}  & \Checkmark{}  & \Checkmark{}  &   &   & \Checkmark{}  & 764  & 21\tabularnewline
\midrule
\Checkmark{}  & \Checkmark{}  & \Checkmark{} & \Checkmark{}  &   &   & 785  & \tabularnewline
\bottomrule
\end{tabular}}

\caption{\label{tab:Single-representation-contributi}Single-representation
contribution in multi-representation setting. Each row shows the setting
of maximal recall when removing one of the four most-contributing
factors (corresponding to the four first columns). Final row shows
performance of all features together. The best performing features
are a combination of those trained for image retrieval by appearance
and shape (\textbf{ret},\textbf{ shp}), alignment of same-class images
(\textbf{weaka}) and categorization (\textbf{conv}). Not shown are
columns corresponding to features not participating in these top few
places. Please refer to text for details. The last column is the reduction
in recall caused by removing the corresponding factor.}
\end{table}

\begin{figure}
\includegraphics[width=1\columnwidth]{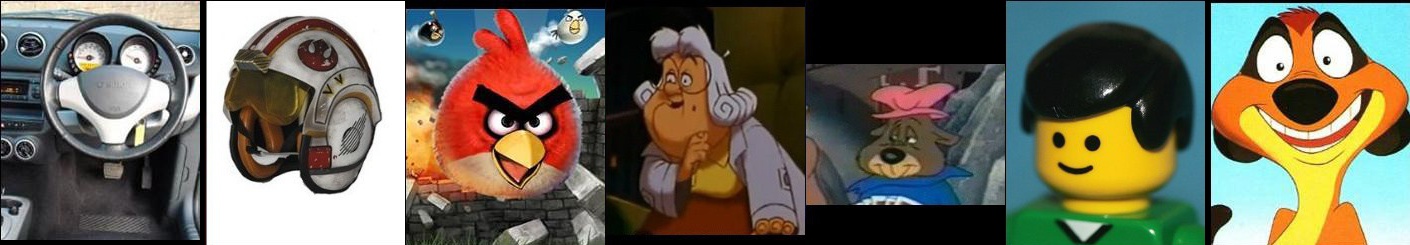}

\includegraphics[width=1\columnwidth]{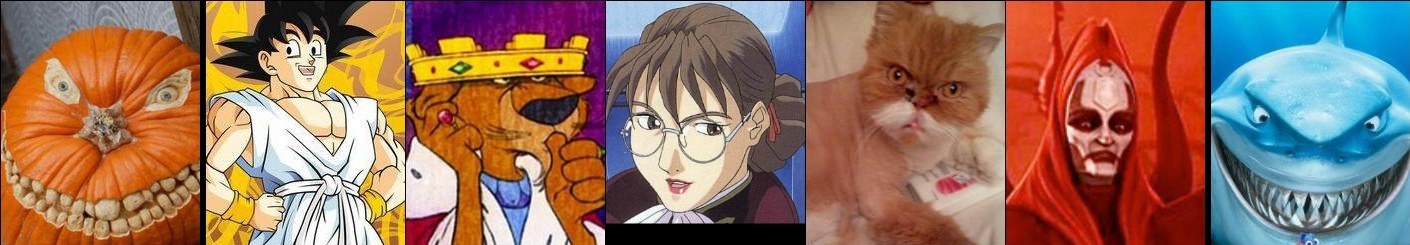}

\includegraphics[width=1\columnwidth]{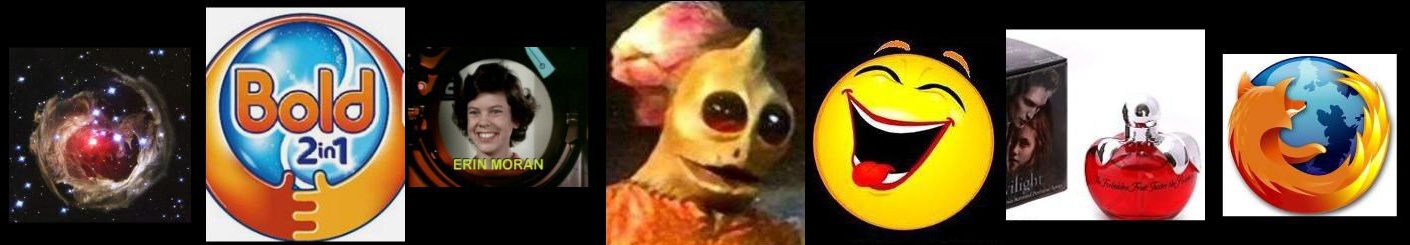}\caption{\label{fig:Failure-cases-on}Failure cases on TLL when using the combined
representation (Section \ref{subsec:Testing-Combined-Representations}).
In each row the first column is the query image, the next five are
the top retrieved ones and the last is the ground-truth.}
\end{figure}

\subsubsection{Feature Specialization \label{subsec:Feature-Specialization}}

\begin{figure*}
\centering{}\subfloat[conv]{\includegraphics[width=0.25\columnwidth]{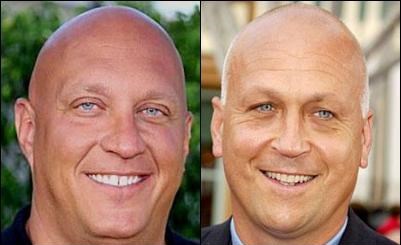}

}\subfloat[pl18]{\includegraphics[width=0.25\columnwidth]{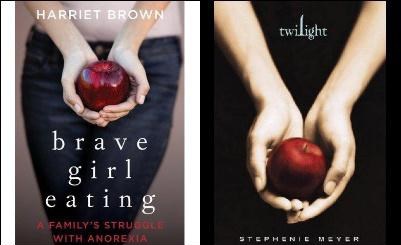}

}\subfloat[pose]{\includegraphics[width=0.25\columnwidth]{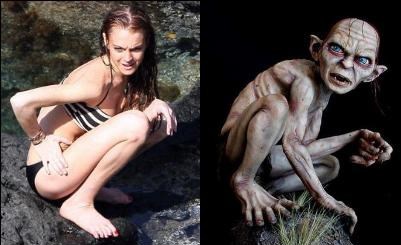}

}\subfloat[ret]{\includegraphics[width=0.25\columnwidth]{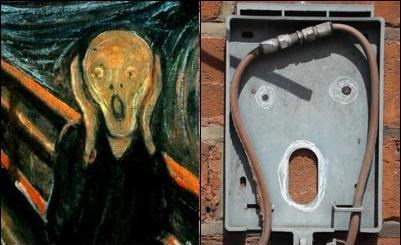}

}\subfloat[ret\_w]{\includegraphics[width=0.25\columnwidth]{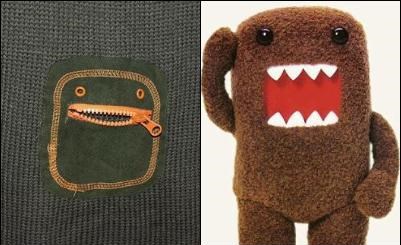}

}\subfloat[shp]{\includegraphics[width=0.25\columnwidth]{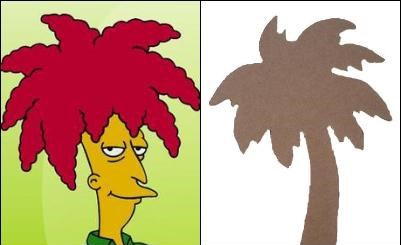}

}\subfloat[texture\_SII]{\includegraphics[width=0.25\columnwidth]{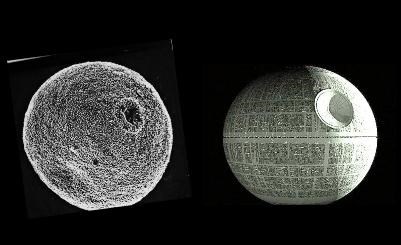}

}\subfloat[weaka]{\includegraphics[width=0.25\columnwidth]{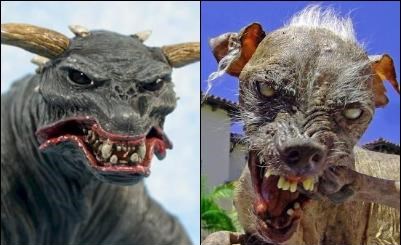}

}\caption{\label{fig:unique-contrib}Exclusively retrieved images. Each image
pair (a-h) show a query image (left) and a successfully retrieved
target image, which was retrieved only using a single feature representation,
and all other representations failed to do so. The diversity of representations
used allows a significant boost in retrieval. Interesting aspects
captured by the representations are those of scene elements (b), pose
(while discarding appearance) (c), local features and layout (d,e),
shape of outline (f), and semantic part similarity (h).}
\end{figure*}

As the various used features seem to be of complementary nature, it
is expected that they will have advantages on different types of images.
This is already shown by the boost in performance above. In this section,
we analyze the contributions of different feature types. Specifically,
for each feature we seek images that were correctly retrieved only
by this feature but not by others. We visualize this in Figure \ref{fig:Feature-importance-vs.}.
Consistent with the results in Table \ref{tab:Single-representation-contributi},
there are relatively many images which are successfully retrieved
by a single representation. In fact, if we allow an ``Oracle'' to
choose the correct representation of each image, the retrieval performance
jumps from 785 to 1073. The overall number of images which are correctly
retrieved exclusively by one feature type is 612. Figure \ref{fig:unique-contrib}
shows some interesting examples of exclusively retrieved images. The
various types of features captured include shape, pose, layout, semantic
part similarity and more. Failure cases of the leading combination
from Section \ref{subsec:Testing-Combined-Representations} are shown
in Figure \ref{fig:Failure-cases-on}. Many of the retrieved images
share some visual aspect with the query (color, local features, shape)
but in the ground-truth answer these features seem to be also geometrically
aligned with the query.

\subsection{Matching Distorted Images}

We now turn to compare our method to the recent one of Zhang \etal
\cite{zhang2018unreasonable}. This work introduces the BAPPS (Berkeley
Adobe Perceptual Patch Similarity) dataset. Each one of over 161k
image patches are distorted in various ways, creating 3-tuples of
\emph{\textless ref}, $p_{0}$, $p_{1}$\textgreater , where \emph{ref
}is a reference (undistorted) patch and $p_{0}$, $p_{1}$ are two
distorted versions. Humans are presented (via Amazon Mechanical Turk)
the reference along with the distorted versions $p_{0}$, $p_{1}$
and the task is to choose which of $p_{0}$, $p_{1}$ seem more similar
to the reference. Two answers are collected for each unique 3-tuple
in their validation set and five for each in the validation set. The
average preference of humans is recorded. The patches are distorted
via several different methods, ranging from traditional ones such
as contrast and saturation modifications to CNN-based ones such as
auto-encoding, denoising, and colorization. We refer the reader to
\cite{zhang2018unreasonable} for more details. Overall, the distortions
tend to modify low-level properties of the images. Their main claim
is that representation in CNNs is can serve as a good approximation
for human perceptual similarity judgments when such low-level distortions
are the main cause of variation. 

We test multiple feature combinations on this dataset. We use the
publicly available code and models from \cite{zhang2018unreasonable}
(the results we report use version 0.1 of the code, and as such may
differ slightly from those in their paper), who use SqueezeNet \cite{iandola2016squeezenet},
AlexNet \cite{krizhevsky2012imagenet}, and VGG16 \cite{simonyan2014very}
networks. To that end, they employ either pretrained versions or the
same architectures trained to predict the human judgments. Their distance
metric is computed as follows. First, they select a subset of layers
from a network. For each layer $l$ the activations are denoted by
$y^{l}\in\mathbb{R}^{H_{l}\times W_{l}\times C_{l}}$. This subset
is fixed throughout the experiments. For example, for AlexNet they
use the ReLu activations after each of the five convolutional layers.
A feature stack is extracted and all activations $y_{hw}$ are unit-normalized
in the channel dimension. For a reference image path $x$ and a reference
$x_{0}$ the distance is computed as follows:
\begin{equation}
d(x,x_{0})=\sum_{l}\frac{1}{H_{l}W_{l}}\sum_{h,w}\left\Vert \omega_{l}\odot(\hat{y}_{hw}^{l}-\hat{y}_{0hw}^{l})\right\Vert _{2}^{2}\label{eq:zhang-similarity}
\end{equation}
where $H,W$ are the height/width of the $l$'th layer and $\omega_{l}\in\mathbb{R}^{C_{l}}$
is a weighting vector, multiplying the difference in each channel
by a scalar weight. Note that this is different than the distance
computation in the previous section, where (1) average global pooling
is first applied to the feature representation and (2) only a single
layer in the network is used. The spatial sensitivity in eq. \ref{eq:zhang-similarity}
makes sense since the distorted images are in fact a modified version
of the reference (assuming there is no severe geometric transformation
being applied). 

If the set of weights $\omega$ is learned, we say that the resulting
distance metric is linearly calibrated. A suffix -lin is then added
to name of the resulting network, for example alex-lin is the linearly
calibrated version of AlexNet.

The score of a method on an image is the fraction of human annotators
that agreed with the method's output. For example, if a method predicts
that $p0$ is closer to the reference than $p1$ then it will receive
0.6 if 3 out of the 5 annotators agree. The score of a method is averaged
over each set of images.

\subsubsection{Semantic vs Low-level Similarity}

We can now ask, does combining multiple feature representations have
the same effect on this task as it does on the TLL challenge? This
is done similarly to Section \ref{subsec:Testing-Combined-Representations}.
We use the pre-trained networks, as well as the linearly calibrated
ones. We test combinations induced by all subsets and report the performance
attained by each network in isolation as well as that attained by
the best combination.

Contrary to the results on the TLL dataset, multiple feature combinations
lead to little or no improvement (\textless 0.5\%). Often the best
performance persists when using a single network. We suggest that
this is because of the nature of the BAPPS dataset. The dataset deals
mainly with low-level feature distortions of images. As these low-level
features seem to be captured well enough by many architectures involving
stacked convolutional layers, there is not much to be gained by combining
several of these architectures. 

To further test this, we also attempted to use all of the features
detailed in Section \ref{subsec:Feature-Representations} and their
combinations. As the total number of feature combinations is prohibitively
large, we limited their number by first testing all combinations up
to size n=4 and finding representations that do not seem to ever contribute
to the overall score when combined with others on this task. These
include the texture\_cnn based features \cite{geirhos2018imagenet}
as well as those using densenet121 \cite{huang2016densely}. Having
discarded these representations, we proceed to test all the combinations
of the remaining ones. The results are shown in Table \ref{tab:perceptual-zhang}.
For each type of distortion we show the single best performing representation
(as indicated by the column ``multiple representations''), and below
it the best representation using any feature combination. In case
where adding the extra features improve performance, we indicate the
performance without the extra features in parentheses. As is evident
by Table \ref{tab:perceptual-zhang}, we see that adding the extra
feature types has a negligible effect in this case. 

We also tested combining the distance metrics with various normalization
techniques as previously, but this did not improve results in this
case.

\subsubsection{Representations of Single Layers\label{par:Representations-of-Single}}

As multiple feature layers are used by \cite{zhang2018unreasonable}
in their linearly calibrated metric, we proceed to test whether it
is really the case that all layers are equally needed. Instead of
using all layers for the computation of the distance, we test the
effect of using only a single layer. This is done by repeating the
calculation in Eq. \ref{eq:zhang-similarity} except that we choose
a single layer $l$ to use, zeroing out $\omega$ for all values corresponding
to other layers and setting it to 1 for $l$. The layers used for
VGG,AlexNet and SqueezeNet are ReLU-activated outputs of various convolutional
layers in each. For VGG and AlexNet this constitutes of five layers
each. For SqueezeNet the use the last five out of seven layers used
by \cite{zhang2018unreasonable}. Table \ref{tab:Matching-perceptual-similarity-single-layer}
compares the results of using the best single layer vs the results
obtained via the linearly-calibrated version of each corresponding
network. $L_{s}$ indicates the layer whose values we use (all others
are zeroed out), where a value of 4 indicates the last (deepest) layer,
3 the preceding layer, and so on; a value of 0 indicates the first
(shallowest) layer of the five. We denote by $\Delta$ the difference
between the linearly-calibrated version and the single-layer version. 

A few observations on the results of Table \ref{tab:Matching-perceptual-similarity-single-layer}:
first, there is little difference ($\Delta)$ in all cases between
the linearly-calibrated distance metric vs that of the best single
layer. Second, in some cases $\Delta$ is negative, indicating that
in fact choosing a single layer can do better than learning the weights
$\omega$. These can be explained because we in fact report the best
performing layer on the validation set, whereas the linear calibration
was trained on the training set. However, another explanation is that
the training procedure of \cite{zhang2018unreasonable} did not include
and l-1 penalty on the weights; perhaps doing so would result in a
more sparse $\omega$. Third, for all distortions at least one network
attains the best performance by choosing a very early layer (0 or
1), and in frameinterp layer 2 is the best for SqueezeNet. This further
indicates that in essence the judgment of similarity in the BAPPS
dataset could be mostly addressed by low-level features. The feature
representation in AlexNet seems to be the overall most useful for
this task.

\begin{table*}
\noindent\resizebox{\textwidth}{!}{%
\begin{tabular}{lrrrrrrrrrrrrrrrrrr}
\toprule 
distortion & \multicolumn{3}{l}{cnn} & \multicolumn{3}{l}{color} & \multicolumn{3}{l}{deblur} & \multicolumn{3}{l}{frameinterp} & \multicolumn{3}{l}{superres} & \multicolumn{3}{l}{traditional}\tabularnewline
net & alex & squeeze & vgg & alex & squeeze & vgg & alex & squeeze & vgg & alex & squeeze & vgg & alex & squeeze & vgg & alex & squeeze & vgg\tabularnewline
\midrule 
$L_{s}$ & 2 & 1 & 0 & 1 & 1 & 4 & 1 & 1 & 3 & 4 & 2 & 3 & 2 & 0 & 2 & 4 & 1 & 4\tabularnewline
score & 82.94 & 82.49 & 81.67 & 64.73 & 63.34 & 60.84 & 60.87 & 60.56 & 59.06 & 62.71 & 62.42 & 62.48 & 71.64 & 71.10 & 69.51 & 73.29 & 76.09 & 75.03\tabularnewline
$\Delta$ & 0.43 & 0.72 & 0.53 & 0.75 & 1.71 & 0.78 & -0.01 & -0.04 & 0.38 & 0.24 & 0.64 & -0.14 & -0.30 & -0.43 & -0.01 & 1.33 & 0.99 & -1.67\tabularnewline
\bottomrule
\end{tabular}

}\caption{\label{tab:Matching-perceptual-similarity-single-layer}Matching perceptual
similarity on the BAPPS dataset \cite{zhang2018unreasonable}. Metrics
computed with a single layer with no per-channel calibration can usually
match the (learned) linearly calibrated metric using all layers. The
layers ($L_{s})$ tend to be early ones, usually several layers from
the last convolutional one. $\Delta$ The absolute difference in performance
vs the linearly calibrated version of the corresponding network. A
negative $\Delta$ means that the learned combination was \emph{worse
}than simply selecting one of the layers.}
\end{table*}

\begin{table*}
\centering{}\noindent\resizebox{\textwidth}{!}{%
\begin{tabular}{c>{\raggedright}p{2.2cm}|>{\centering}p{1cm}|>{\centering}p{0.7cm}|c|c|c|c|c|c|c|c|c|c|c|c|>{\centering}p{1cm}|c|}
\hline 
 & \multirow{1}{2.2cm}{multiple representations?} & alexnet$\dagger$ & pose & ret & vgg16 & weaka & shp & retw & alex-lin & squeeze & alex & vgg-lin & L2 & vgg$\dagger$ & SSIM & squeeze-lin & score\tabularnewline
\hline 
cnn & no &  &  &  &  &  &  &  & \Checkmark{} &  &  &  &  &  &  &  & 83.37\tabularnewline
 & yes &  &  &  &  &  &  &  & \Checkmark{} &  &  &  &  &  &  & \Checkmark{} & \textbf{83.68}\tabularnewline
\hline 
color & no &  &  &  &  &  &  &  & \Checkmark{} &  &  &  &  &  &  &  & 65.47\tabularnewline
 & yes &  &  &  &  &  &  &  & \Checkmark{} &  &  &  & \Checkmark{} &  &  & \Checkmark{} & \textbf{65.87}\tabularnewline
\hline 
deblur & no &  &  &  &  &  &  &  & \Checkmark{} &  &  &  &  &  &  &  & 60.87\tabularnewline
 & yes &  &  &  &  &  &  &  & \Checkmark{} &  &  &  & \Checkmark{} &  &  & \Checkmark{} & \textbf{61.10}\tabularnewline
\hline 
frameinterp & no &  &  &  &  &  &  &  &  &  &  &  &  &  &  & \Checkmark{} & 63.05\tabularnewline
 & yes &  &  &  &  &  &  &  & \Checkmark{} &  &  &  &  &  &  & \Checkmark{} & \textbf{63.72}\tabularnewline
\hline 
superres & no &  &  &  &  &  &  &  &  &  & \Checkmark{} &  &  &  &  &  & 71.66\tabularnewline
 & yes & \Checkmark{} &  &  &  & \Checkmark{} &  &  & \Checkmark{} &  & \Checkmark{} &  &  &  & \Checkmark{} & \Checkmark{} & \textbf{71.81 }(71.77)\tabularnewline
\hline 
traditional & no &  &  &  &  &  &  &  &  &  &  &  &  &  &  & \Checkmark{} & 77.08\tabularnewline
 & yes &  &  &  & \Checkmark{} &  & \Checkmark{} &  & \Checkmark{} &  &  &  &  &  &  & \Checkmark{} & \textbf{77.66 }(77.08)\tabularnewline
\hline 
\end{tabular}}\caption{\label{tab:perceptual-zhang}Multiple feature combinations for perceptual
similarity on BAPPS dataset. Combining multiple features does not
seem to gain anything significant for this data.$\dagger$ uses the
last convolutional layer with global average pooling before the distance
computation. -lin suffix: linearly calibrated feature computations.
Scores in parentheses () show results without the extra features,
if the extra features resulted in some improvement.}
\end{table*}

\section{Discussion \& Conclusions}

We have examined a method of combining multiple feature representations
in order to reproduce human perceptual similarity judgments. The two
datasets on which we tested exhibit markedly different properties.
The first, TLL \cite{rosenfeld2018totally} presents similarities
that stem from various high-level features in the images, including
shape, pose and faces. The second, of a much larger scale (BAPPS \cite{zhang2018unreasonable})
exhibits low-level differences between the images. Our examination
reveals a few differences between the two. The similarity judgments
in BAPPS seem to be predicted quite well by using low level features
readily available in various layers in convolutional neural networks.
Attempting to learn these similarities does not seem to gain much
in terms of prediction accuracy \cite{zhang2018unreasonable}. In
fact, some of the learned versions perform even worse than simply
using a single layer within the networks, and often an early layer
(Section \ref{par:Representations-of-Single}). Combining multiple
representations from various sources does not improve these results.

On the TLL dataset, any single representation does quite poorly in
predicting the similarity judgments. However, combining various representations
has a dramatic improvement over the results, indicating that the features
required for this task indeed span multiple types or domains. Contrary
to the current common wisdom that a stronger baseline for one task
will perform better on another, the generalization power of various
networks on a single task (ImageNet \cite{krizhevsky2012imagenet})
does not seem to be a good indication of its success on the TLL benchmark
(Figure \ref{fig:Performance-of-each}). Training a method to predict
similarity judgments solely on a dataset such as TLL seems like a
difficult task. One reason is that TLL may be prohibitively small
(\httilde6000 images) with respect to its variability. One option
is to extend its size and diversity, as is the common trend. Our results
suggest a complementary approach: human similarity judgments are multi-faceted,
involving various feature types of different semantic levels. Moreover,
the networks whose combined representations improve the performance
are more easy to train. Classification data has already been collected
at large scale \cite{russakovsky2015imagenet} and is relatively easy
for humans to annotate (at the image level). For image retrieval clever
methods have been devised to use classical vision methods \cite{radenovic2018fine}
to collect training data. Using these networks and others results
in a powerful representation for a task which none of them do particularly
well. In this, we provide a complementary view to the work of Zhang
\etal \cite{zhang2018unreasonable}, who show that representations
suitable for perceptual similarity judgments result from training
deep convolutional neural networks, regardless of the training signal,
as long as one exists (\ie untrained networks perform significantly
worse in this regard). Their BAPPS dataset consists of low level image
distortions. We believe that the computational prior in conv-nets
is hence responsible for their findings. Indeed, strongly trained
networks ($\eg$, on ImageNet) perform on the BAPPS dataset almost
as well as those trained specifically on the dataset itself. In our
work, we find that for high-level, semantic similarity judgments,
one needs to employ diverse sources; low level priors no longer suffice.
Moreover, the same networks that perform well on BAPPS do not perform
well at all on TLL - and vice versa: adding high level networks that
contributed to the performance of TLL gained very little for BAPPS.

Though the results are encouraging, they are far from perfect. In
future work, we intend to explore how additional useful representations
may arise by solving a variety of tasks. This can aid in tasks who
are not practical to directly supervise at scale or even improve those
who are. It can also help explore how representations emerge in humans
without explicit supervision.

\bibliographystyle{unsrt}
\bibliography{egbib}

\end{document}